\pdfoutput=1

\documentclass[11pt]{article}

\usepackage[final]{acl}

\usepackage{times}
\usepackage{latexsym}

\usepackage[T1]{fontenc}

\usepackage[utf8]{inputenc}

\usepackage{microtype}

\usepackage{inconsolata}

\usepackage{graphicx}

\usepackage[utf8]{inputenc} 
\usepackage[T1]{fontenc}    
\usepackage{hyperref}       
\usepackage{url}            
\usepackage{booktabs}       
\usepackage{amsfonts}       
\usepackage{nicefrac}       
\usepackage{microtype}      
\usepackage{xcolor}         

\usepackage{algorithmic}
\usepackage{algorithm}

\usepackage{graphicx}
\usepackage{caption}

\usepackage{amsmath}
\usepackage{amssymb}
\usepackage{mathtools}
\usepackage{amsthm}
\usepackage{array}
\usepackage{adjustbox}

\usepackage{float}
\usepackage{tabularx} 
\usepackage{subcaption}
\usepackage{multirow}    
\usepackage{tabularray}

\usepackage{hyperref}       
\usepackage{csquotes}
\usepackage{csvsimple} 
\usepackage{dirtytalk} 
\usepackage{cite}
\usepackage{pgf-pie} 
\definecolor{darkab52c5}{rgb}{0.82, 0.38, 0.89} %
\definecolor{customblue}{HTML}{0a29cb}
\definecolor{custompurple}{HTML}{ab52c5}
\usepackage{lineno}
\definecolor{darkblue}{rgb}{0, 0, 0.5}
\hypersetup{colorlinks=true, citecolor=darkblue, linkcolor=darkblue, urlcolor=darkblue}

\usepackage[capitalize,noabbrev]{cleveref}

\title{Leveraging NTPs for Efficient Hallucination Detection in VLMs}

\author{
 \textbf{Ofir Azachi\thanks{Equal contribution.}\textsuperscript{,1}},
 \textbf{Kfir Eliyahu\footnotemark[1]\textsuperscript{,1}},
 \textbf{Eyal El Ani\footnotemark[1]\textsuperscript{,1}},
 \textbf{Rom Himelstein\footnotemark[1]\textsuperscript{,1}},\\
  \textbf{Roi Reichart\textsuperscript{1}},
\textbf{Yuval Pinter\textsuperscript{2}},
\textbf{Nitay Calderon\textsuperscript{1}}
\\
 \textsuperscript{1}Department of Data and Decision Science, Technion - Israel Institute of Technology, \\
 \textsuperscript{2} Faculty of Computer and Information Science, Ben-Gurion University of the Negev
 \\
 \small{
   \textbf{Correspondence:} \href{mailto:romh@campus.technion.ac.il}{romh@campus.technion.ac.il}.
 }
}

\begin{document}
\maketitle

\begin{abstract}
Hallucinations of vision-language models (VLMs), which are misalignments between visual content and generated text, undermine the reliability of VLMs. One common approach for detecting them employs the same VLM, or a different one, to assess generated outputs. This process is computationally intensive and increases model latency. In this paper, we explore an efficient on-the-fly method for hallucination detection by training traditional ML models over signals based on the VLM's next-token probabilities (NTPs). NTPs provide a direct quantification of model uncertainty. We hypothesize that high uncertainty (i.e., a low NTP value) is strongly associated with hallucinations. To test this, we introduce a dataset of 1,400 human-annotated statements derived from VLM-generated content, each labeled as hallucinated or not, and use it to test our NTP-based lightweight method. Our results demonstrate that NTP-based features are valuable predictors of hallucinations, enabling fast and simple ML models to achieve performance comparable to that of strong VLMs. Furthermore, augmenting these NTPs with linguistic NTPs, computed by feeding only the generated text back into the VLM, enhances hallucination detection performance. Finally, integrating hallucination prediction scores from VLMs into the NTP-based models led to better performance than using either VLMs or NTPs alone. We hope this study paves the way for simple, lightweight solutions that enhance the reliability of VLMs. All data is publicly available at \href{https://huggingface.co/datasets/wrom/Language-Vision-Hallucinations}{\includegraphics[height=1em]{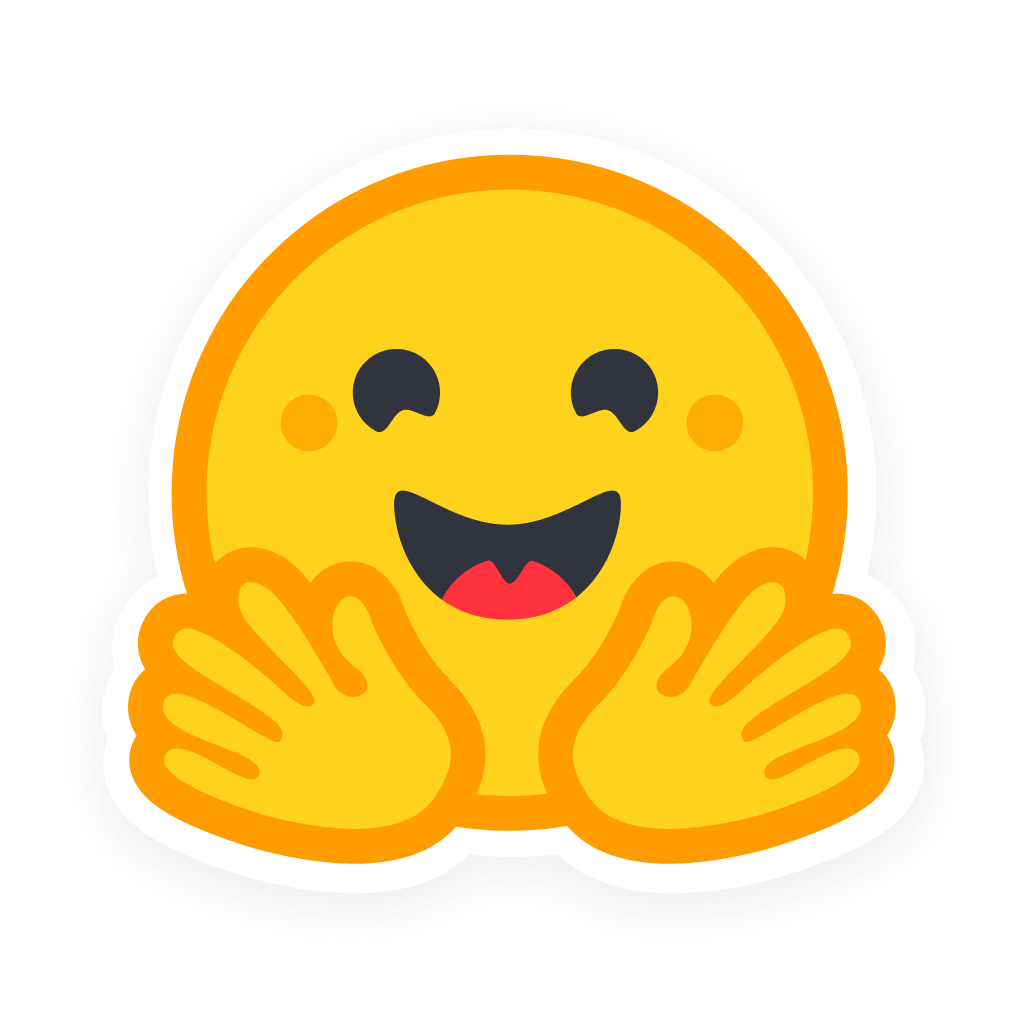}}.

\end{abstract}

\section{Introduction}

\emph{Vision-language models (VLMs)} have emerged as powerful tools capable of handling tasks involving visual and textual inputs. These models enable applications such as visual question answering~\citep[VQA;][]{li2019visualbertsimpleperformantbaseline}, and text-to-image generation~\citep{radford2021learningtransferablevisualmodels, zhao2024evolvedirector}. However, as these models become more widely used, concerns about \emph{hallucinations}, errors or misleading outputs generated by the model, have become more prominent.
Unlike humans, who are less likely to describe non-existent objects, misjudge colors, or miscount elements, these errors are more likely to appear in machine-generated content.
\citet{gunjal2024detecting} found that even state-of-the-art VLMs frequently generate non-existent objects.

\begin{figure}[t]
    \centering
    \includegraphics[width=\linewidth]{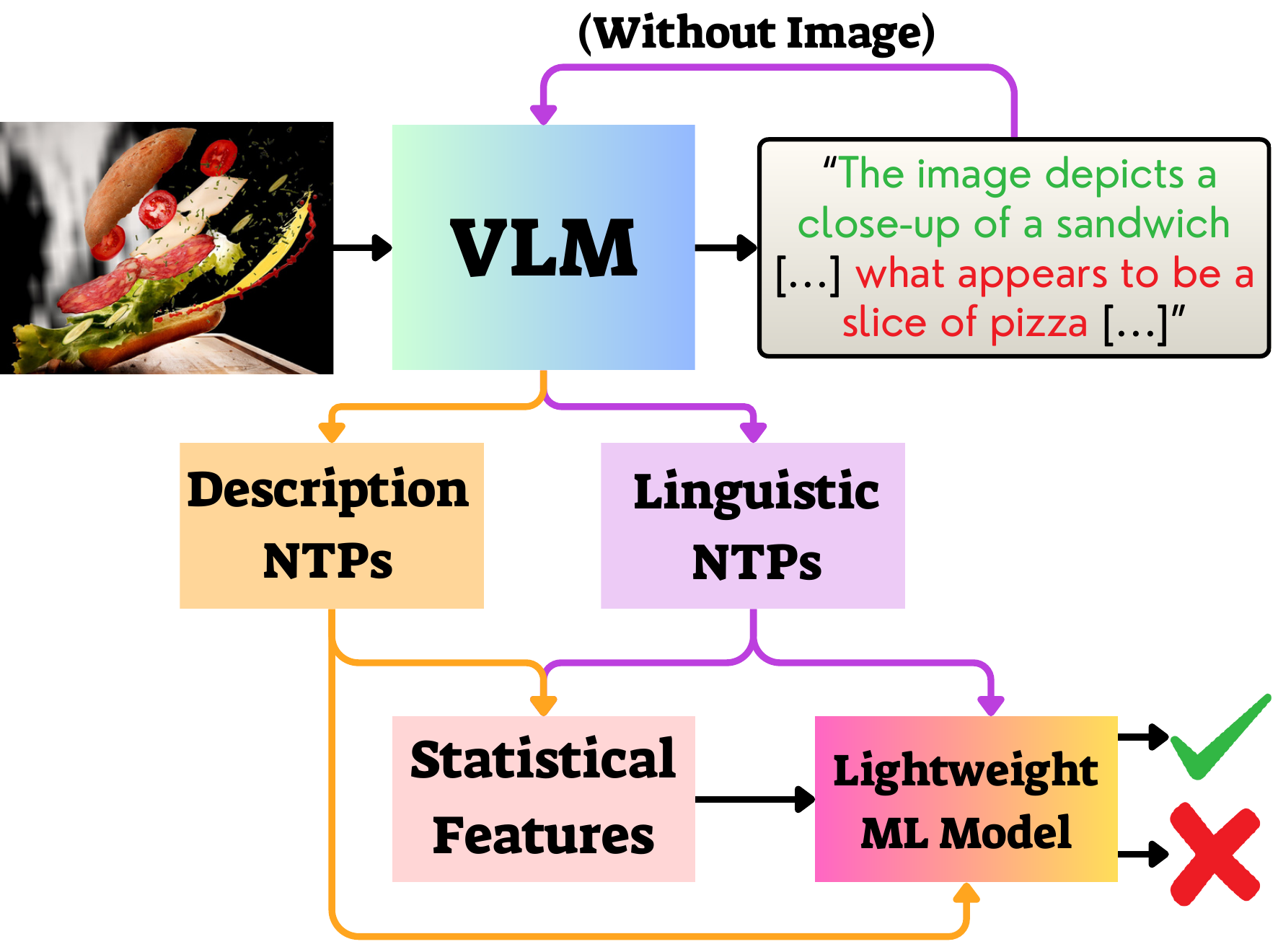}
    \caption{\textbf{Illustration of our method:} Linguistic NTPs are extracted during the VLM’s text generation process. Description NTPs require an additional forward pass using only the generated text. Statistical features are then computed from the NTPs, and a lightweight traditional ML model uses these features to detect hallucinations.}
    \label{fig:teaser}
    \vspace{-0.8em}
\end{figure}

Currently, the primary method for detecting hallucinations involves using VLMs as hallucination predictors, either by asking a model to identify hallucinations in its own generated output or in others'~\citep{li2024reference}.
This approach has demonstrated success both in generative LLMs~\citep{quevedo2024detecting} and generative VLMs~\citep{chen2024unified}.
However, these predictor VLMs exhibit two main weaknesses: First, they often require performing extensive computations, making them both computationally expensive and time-consuming, especially when multiple calls are needed to verify each sentence or clause in the generated content.
Second, they lack explainability and interpretability~\citep{zhao2024explainability}. 

\emph{Large language models (LLMs)} generate responses by sampling tokens from a learned probability distribution over the next token, conditioned on the input context. This auto-regressive generation process resembles human language production, where likely words are uttered based on contextual understanding and prior knowledge~\citep{goldstein2022shared}. \citet{lu2021communication} found that, in humans, uncertainty plays a key role in the propagation of misinformation. Inspired by this, we hypothesize that \emph{next-token probabilities (NTPs)} produced by VLMs may similarly encode uncertainty, and thus can serve as useful signals for hallucination detection. Indeed, prior work suggests that high uncertainty, reflected by low NTPs, is a strong indicator of hallucinations and related errors~\citep{farquhar2024detecting,quevedo2024detecting,li2024reference}.

We investigate the role of NTPs in detecting hallucinations in VLMs. Rather than relying on predictor VLMs, we propose leveraging the NTPs produced during generation to enable fast, real-time hallucination detection. Our goal is to design an effective approach for leveraging NTP-based features to predict hallucinations using fast, lightweight traditional machine learning (ML) models, such as Logistic Regression, Support Vector Machine, and XGBoost. As illustrated in Figure~\ref{fig:teaser}, we compare approaches that use raw NTPs directly from the VLMs (\emph{Description NTPs}) with those that rely on statistical features derived from the NTPs. We explore integration of NTPs with VLM predictor outputs, and propose a method for neutralizing linguistic biases embedded within them using \emph{Linguistic NTPs} resulting from reprocessing the generated text through the same VLM after omitting the visual input. Throughout, we assume that higher uncertainty, operationalized as lower next-token probabilities or higher entropy, correlates with hallucination risk~\citep{farquhar2024detecting}, and we design our features to capture this signal.

A growing body of research shows that VLMs often rely heavily on linguistic priors~\citep{zhu2024ibd,guan2024hallusionbench,cho2025influence,wang2024linguistic}, and may even prioritize them over conflicting visual evidence~\citep{luo2024vvlm,wu2024autohallusion}. These findings suggest that hallucinations in VLMs may stem, at least in part, from biases in their language modeling components, rather than solely from limitations in visual understanding.
Based on these insights, we introduce a novel dataset specifically curated to examine the relationship between NTPs and hallucinations in VLMs.
We believe this dataset will serve as a valuable resource for future research in this area. Using this dataset, we evaluate the effectiveness of NTPs generated by \emph{LLaVA-1.5} and \emph{LLaVA-1.6} \citep{liu2024visual} for hallucination detection. As baselines, we include predictions from both \emph{LLaVA} and \emph{PaliGemma} \citep{PaliGemma}, and also use these predictions as additional input features to traditional ML models. 

Our experiments reveal that statistical features derived from NTPs outperform raw NTP features across all models, making them a more effective and reliable signal for hallucination detection. These statistical features alone come close to matching the performance of VLM predictors while offering gains in efficiency, allowing for on-the-fly hallucination detection. While incorporating \emph{Linguistic NTPs} offers only modest gains for statistical features, neutralization strategies such as element-wise subtraction of raw \emph{Description} and \emph{Linguistic} NTPs provide further evidence of the role of linguistic biases in hallucination generation. Finally, we find that augmenting VLM predictor outputs with NTP features yields consistent improvements, demonstrating that these signals are complementary and result in the strongest hallucination detection approach.






\section{Related work}

\textbf{Defining hallucinations.}
The term \textit{hallucinations} lacks a universal definition across different fields but, in general, describes instances where a model produces content that is disconnected from its input or from reality~\citep{maleki2024ai}.
In NLP, this term typically refers to outputs that fail to accurately reflect real-world facts~\citep{xu2024hallucination}.
The notion extends to other areas as well; for example, in medical imaging, deep learning techniques can create images that appear realistic, but contain fabricated structures, potentially misleading diagnostic efforts~\citep{bhadra2021hallucinations}. Identifying hallucinations is critical because inaccuracies not only diminish user trust but also present significant risks across diverse domains~\citep{benkirane2024machine, tang2025mitigatinghallucinatedtranslationslarge}, including low-resource language settings~\citep{benkirane2024machine}, legal contexts~\citep{magesh2024hallucination}, information retrieval~\citep{faggioli2023perspectives}, healthcare and autonomous driving~\citep{leng2024mitigating, gunjal2024detecting}. Consequently, robust hallucination detection is essential to mitigate these challenges and safeguard the reliability of AI-generated content.


\textbf{Techniques for hallucination detection.} Various methods have been proposed to automatically detect hallucinated outputs.
One common approach involves analyzing the model's output probability distributions, where segments with low confidence, characterized by high entropy or significantly reduced token probabilities, are reliably flagged as hallucinations~\citep{li2024reference,ma2025estimating,guerreiro2022looking,quevedo2024detecting,farquhar2024detecting,simhi2025trust}.
In contrast to these internal indicators, other methods deploy external models such as dedicated VLMs~\citep{chen2024unified} or LLMs~\citep{quevedo2024detecting} to assess whether hallucinations are present in the generated content.
Although this external verification yields promising results, it tends to be significantly more resource-intensive than relying solely on internal signals, and lacks explainability~\citep{sarkar2024large,zhao2024explainability}.


\textbf{Linguistic biases and their impact on VLMs.}
A significant source of hallucinations in both VLMs and LLMs is their overdependence on linguistic priors and biases.
Research indicates that large VLMs often generate plausible-sounding descriptions based on statistical patterns learned during training (e.g., "blue sky"), rather than by accurately anchoring every detail to the visual content~\citep{zhu2024ibd, guan2024hallusionbench}.
This can result in errors such as attributing objects or attributes to a scene that, while contextually expected, are actually absent—a phenomenon commonly known as \emph{object hallucination} in image captioning and VQA systems~\citep{leng2024mitigating}.
In many cases, the language generation component can dominate the visual signal, with models relying solely on textual context even when it contradicts the visual evidence~\citep{luo2024vvlm,wu2024autohallusion}.
Consequently, recent research focuses on minimizing these linguistic biases to reduce hallucinations originating from the multimodal interaction, for instance, by encouraging the model to more closely attend to the image during the decoding process~\citep{zhu2024ibd,leng2024mitigating}.

\section{Method}
\label{sec:method}

\textbf{Problem definition.}
A \emph{probe} is a statement derived from a VLM-generated description of an image. Each probe can either be truthful or contain a hallucination. For example, the probe \textit{`There is a handbag.'} from \autoref{fig:data_sample} corresponds to the generated sentence \textit{`There is also a handbag visible in the scene.'} We define \emph{hallucinations} as any textual information produced by the VLM that does not accurately reflect the visual content of the image. In particular, we consider the following as hallucinations: objects falsely perceived as present, incorrect object attributes (such as color or size), and misinterpretations of relationships within the scene. Our goal is to predict whether a probe contains a hallucination or not.

\subsection{Predicting Hallucinations}
\label{subsec:predict_hallucination}

We employ two complementary approaches to predict whether a probe contains a hallucination.
The first approach employs a predictor VLM (e.g., \emph{LLaVA-1.5}, \emph{LLaVA-1.6} or \emph{PaliGemma}) which process the image using the prompt:
\begin{quote}
``According to the image, is the following sentence correct? \{PROBE\}. Answer only with Yes OR No.''
\end{quote}
Here, \{PROBE\} represents a probe derived from the VLM-generated description of the image. We denote the probability that the probe is correct, as estimated using the NTP of the predictor VLM, by:
\[
\frac{\mathbb{P}(\text{Yes})}{\mathbb{P}(\text{Yes}) + \mathbb{P}(\text{No})}
\]

The main drawback of this approach is the reliance on a predictor VLM, which can be computationally expensive. In real-time applications, where we aim to verify that the content generated by the VLM is correct, this approach substantially increases latency, as each statement is verified separately. To address this, we propose an alternative approach that employs fast and lightweight traditional machine learning models (we use the term \textit{traditional ML models} in the remaining of the text), such as Logistic Regression (LR), Support Vector Machine (SVM), and XGBoost. These models are trained to predict whether a probe is correct based on features derived from the NTPs of the VLM-generated description. Since these NTPs are by-products of the generation process, the models can assess the correctness of the generated content on the fly (i.e.,~during generation). In the following subsection, we describe these NTP-based features.

\subsection{Next Token Probabilities (NTPs)}
\label{subsec:ntp_representations}

We present two types of NTPs that are used as features for the traditional ML models.

\textbf{Description NTPs.} When a VLM generates a response, it does so token by token, estimating a probability distribution over all possible tokens at each step. We hypothesize that these \emph{Description NTPs} encode valuable information about the model's certainty in its generated response and, therefore, may be beneficial for hallucination detection. Since \emph{Description NTPs} can be obtained on the fly, they serve as our primary focus.

\textbf{Linguistic NTPs.} Our manual analysis of \emph{Description NTPs} revealed recurring probability patterns that suggest linguistic influences beyond visual content. We hypothesize that these patterns arise from inherent linguistic biases in the model. Following the methodology of \citet{Liu2023TemperaturescalingSE,Shrivastava2023LlamasKW}, who demonstrated that linguistic effects in the generated text could be captured by feeding the text back into the same language model that produced it, we reinserted the VLM-generated text into its corresponding language model, this time without an instructional prompt or image. We term the extracted probabilities as \emph{Linguistic NTPs}. Our motivation is to augment the \emph{Description NTPs} with \emph{Linguistic NTPs}, which help disentangle language-driven biases, such as syntactic or grammatical priors, from visually grounded signals, thereby improving the detection of hallucinated content.

To quantify the relationship between \emph{Description NTPs} and \emph{Linguistic NTPs}, we computed Spearman's correlation between the two probability series for each probe. The average correlation across all probes was $0.744$, reinforcing our hypothesis that the two types of NTPs are inherently linked. Consequently, we examine the potential of \emph{Description NTPs} both as standalone features and in combination with \emph{Linguistic NTPs}.

\subsection{Next Token Probabilities as Features}
\label{subsubsec:padded}

We next describe how the NTPs are used in practice as features for traditional ML models. \emph{Description NTPs} are extracted on the fly during text generation. For each probe, we consider only the NTPs corresponding to the span of generated text associated with that probe, typically a sentence or clause, though not necessarily limited to that. \emph{Linguistic NTPs}, on the other hand, are extracted separately, either after the full description has been generated or after the span corresponding to each probe (e.g., after each sentence). The result is one (or two) matrices with a shape equal to the number of generated tokens in the span by the vocabulary size. In our main setup, we use only the probability values assigned by the VLM to the actually generated tokens, resulting in a dense vector of length equal to the number of tokens in the span. 

Naturally, using these vectors as raw features presents several challenges. First, spans may vary in length, whereas traditional ML models require a fixed number of input features. Second, there are multiple ways to combine the \emph{Description} and \emph{Linguistic NTPs}. Third, the sequences can be long, which motivates aggregation and feature engineering. To address the challenge of varying sequence lengths, each sequence of NTPs (either \emph{Description} or \emph{Linguistic}) is zero-padded to match the length of the longest sequence in the dataset, which contains 42 tokens. To explore how to best combine the two types of NTPs, the following aggregation techniques were applied:

\begin{figure*}[t]
    \centering
    \includegraphics[width=1\textwidth]{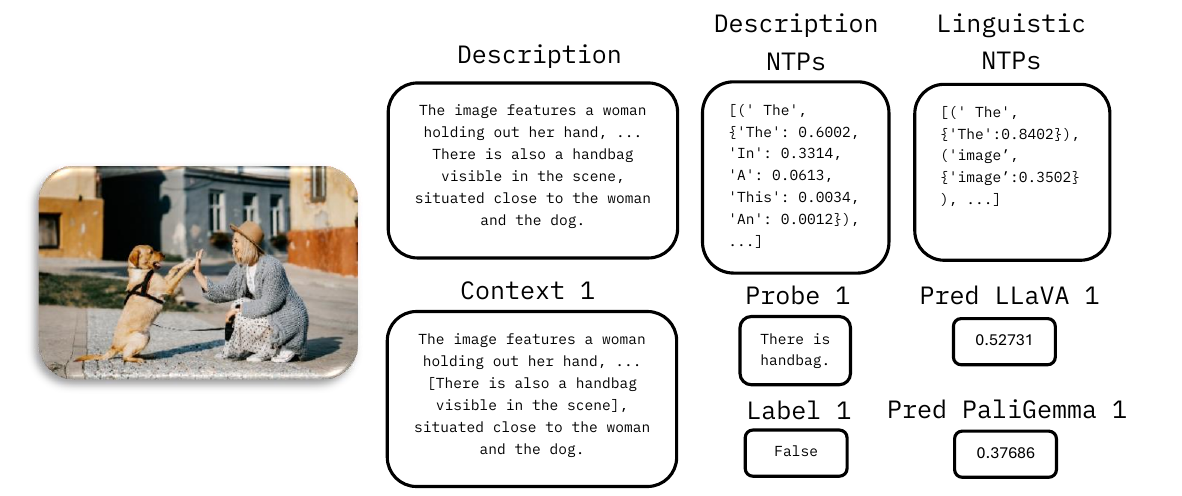}
            \vspace{-0.8em}
    \caption{An example for the data features.}
    \label{fig:data_sample}  
\end{figure*}

\begin{itemize}
\item \textbf{Only Description NTPs:} Use only the  \emph{Description NTPs} as input features.
\item \textbf{Only Linguistic NTPs:} Use only the \emph{Linguistic NTPs} as input features.
\item \textbf{Concatenation:} Concatenate the \emph{Description} and \emph{Linguistic NTPs} sequences, resulting in a combined input of 84 features.
\item \textbf{Element-wise subtraction:} Subtract the \emph{Linguistic NTPs} from the \emph{Description NTPs} token by token.
\item \textbf{Element-wise division:} Divide the \emph{Description NTPs} by the \emph{Linguistic NTPs} token by token using:
\[
t^{\text{div}}_i = \frac{t^{\text{Desc}}_i}{1 + t^{\text{Ling}}_i} \in [0,1],
\]
where \( t_i \) represents the corresponding NTP value pf the $i$-th generated token.
\end{itemize}
While \textit{raw NTP} values provide direct probabilistic information, they may not capture higher-level patterns or summarise statistics that might be useful for hallucination detection. To enrich the feature space, we also engineer \textit{statistical features}:
\begin{itemize} 
\item Mean of the generated-token NTPs.
\item Standard deviation of the NTPs. 
\item Mean of the logarithm and exponent of the NTPs ($\log(\mathbb{P})$ and $\exp(\mathbb{P})$).
\item The top-$k$ dominant frequencies (excluding DC) from the Discrete Fourier Transform of real-valued NTPs, where $k$ is a hyper-parameter ranging from 0 to 5 (0 serving as the control).
\end{itemize}
If both types of NTPs are available, we extract the following additional features:
\begin{itemize}
\item Mean of the element-wise product between the \emph{Description} and \emph{Linguistic NTPs}.
\item Minimum between (i) the mean of the element-wise ratio of \emph{Linguistic NTPs} to \emph{Description NTPs}, and (ii) the mean of the element-wise ratio of \emph{Description NTPs} to \emph{Linguistic NTPs}.
\end{itemize}

\section{Hallucination Detection Dataset}
\label{sec:data}

Our dataset consists of 350 images, sourced from Pixabay\footnote{\href{https://pixabay.com/}{https://pixabay.com/}} and iStock.\footnote{\href{https://www.istockphoto.com/}{https://www.istockphoto.com/}}
For each image, a \emph{LLaVA} model was prompted with the instruction: \say{Please provide a thorough description of this image}. The generated descriptions were manually reviewed, and only those containing at least one hallucination were retained. This procedure yielded 200 examples using \emph{LLaVA-1.6} and 150 examples using \emph{LLaVA-1.5}. From each VLM-generated description with at least one hallucination, four probes were extracted, ensuring that at least one probe per description contained a hallucination. In total, the dataset comprises 1,400 probes, of which 42.9\% are labeled as hallucinated. The annotation process was conducted by a group of seven undergraduate students (six males and one female), with ages ranging from 21 to 28 years. Each data sample includes the following features, with $i\in[4]$:

\textbf{Description:} The generated description by the \emph{LLaVA} model. \textbf{Description NTPs:} The NTPs of the \emph{LLaVA} generated tokens.\footnote{We also saved non-generated tokens with probabilities above a set minimum threshold of $1\text{e-}3$.} \textbf{Linguistic NTPs:} A sequence of probabilities, where each value represents the likelihood of a generated token when the description is processed without the image input. $\mathbf{Probe (i)}$: A statement written by the annotators that can be derived from the respective Description. At least one probe among the four contains a hallucination. $\mathbf{Label (i)}$: A binary label (True/False) that was manually assigned to decide the validity of $Probe (i)$. $\mathbf{Context (i)}$: A markup of the part of the generated description that $Probe (i)$ refers to from the respective Description. $\mathbf{LLaVA~Pred (i)}$: The \emph{LLaVA} VLM estimation of $Probe (i)$'s correctness, as described in \S\ref{subsec:predict_hallucination}. $\mathbf{PaliGemma~Pred (i)}$: The \emph{PaliGemma} VLM estimation of $Probe (i)$'s correctness, see \S\ref{subsec:predict_hallucination}.


\begin{figure*}[t]
    \centering
    \includegraphics[width=0.9\textwidth]{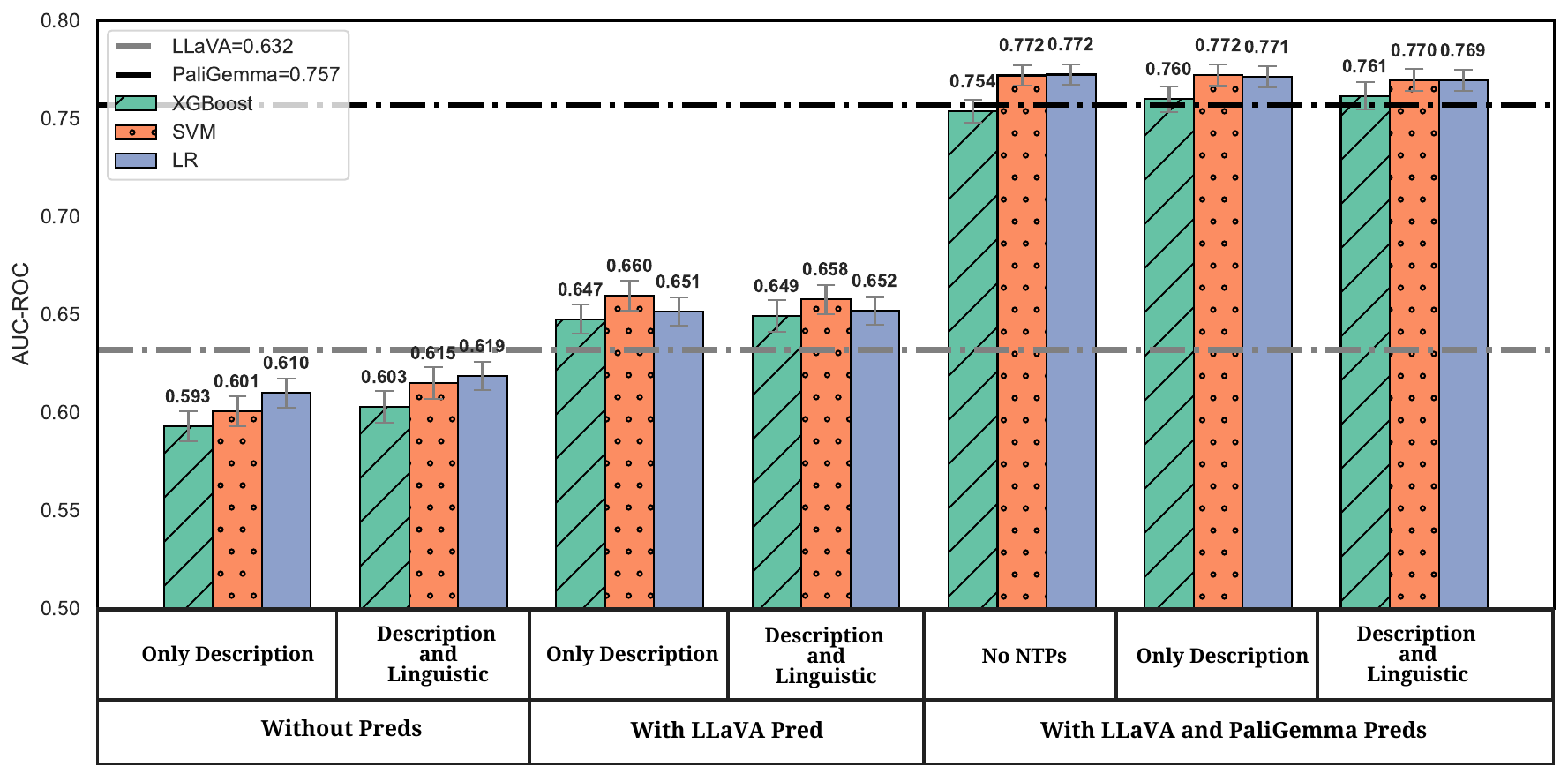}
    \caption{AUC-ROC performance of traditional ML models using statistical features of NTPs and various $\mathbf{Pred}$ features. Each bar group corresponds to a specific feature combination, while the dashed lines denote the \emph{LLaVA} and \emph{PaliGemma} baselines. Error bars indicate 95\% confidence intervals.}
    \label{fig:Engineered_Features_results}
    \vspace{-0.8em}
\end{figure*}

\autoref{fig:data_sample} illustrates the features described above. An example of the data collection pipeline is provided in \autoref{sec: AppendixPipeline}.
A detailed analysis of the \emph{Description NTPs} and \emph{Linguistic NTPs} is presented in \autoref{appendix:motivation}, along with supporting evidence for their potential usefulness as input features to the models introduced in the following section.

\section{Experimental Setup}

\textbf{VLM Predictors} To evaluate the effectiveness of probe-based hallucination detection, we employ two VLM predictors.
The first is a \emph{LLaVA}-based predictor,\footnote{\href{https://huggingface.co/llava-hf/llava-1.5-7b-hf}{huggingface.co/llava-hf/llava-1.5-7b-hf}; \href{https://huggingface.co/llava-hf/llava-v1.6-mistral-7b-hf}{huggingface.co/llava-hf/llava-v1.6-mistral-7b-hf}} corresponding to the same VLM that generated the image description.
The rationale is to compare the performance of traditional ML models that rely on the VLM’s NTPs with that of using the same VLM for self-verification of its own generated content.
The second VLM predictor is an external model, \emph{PaliGemma}.\footnote{\href{https://huggingface.co/google/PaliGemma-3b-pt-224}{huggingface.co/google/PaliGemma-3b-pt-224}}
Naturally, using an external VLM also imposes additional computational and memory overhead.

\textbf{Traditional ML models} We experiment with three traditional ML models: Logistic Regression (LR), Support Vector Machine (SVM), and XGBoost. 
We employ two sets of features, as described in \S\ref{subsubsec:padded}: (i) raw NTPs, using either \emph{Description NTPs}, \emph{Linguistic NTPs}, or a combination of both; and (ii) statistical features extracted from the NTPs. Each model is trained on $1000$ examples ($71.4$\% of the full dataset), with an additional $200$ examples ($14.3$\%) used for validation (for hyperparameter tuning), and evaluated on a test set of $200$ examples ($14.3$\%). To ensure the robustness of our results, the reported results reflect the average performance over $100$ random splits.

\textbf{Combining NTP-based features with VLM predictors} We investigate whether combining the $\mathbf{Pred}$ feature obtained from a predictor VLM (\emph{LLaVA} or \emph{PaliGemma}) with the NTP-based features improves detection.
Accordingly, the input of traditional ML models is augmented with one or both predictor outputs.
While this approach introduces additional computational cost due to extra VLM inference, it allows us to assess whether combining fast NTP-based features with direct VLM predictions offers complementary benefits.

\textbf{Hyperparameter tuning.} We perform hyperparameter tuning for each train-validation split to ensure optimal model performance. The tuning process aims to maximize the Area Under the ROC Curve (AUC-ROC) on the validation set. Given the variability in input representations and model configurations, the specific hyperparameter ranges for each setting are provided in Appendix~\ref{app:hyperparams}.

\section{Results}

\begin{figure*}[ht]
    \centering
    \includegraphics[width=0.9\textwidth]{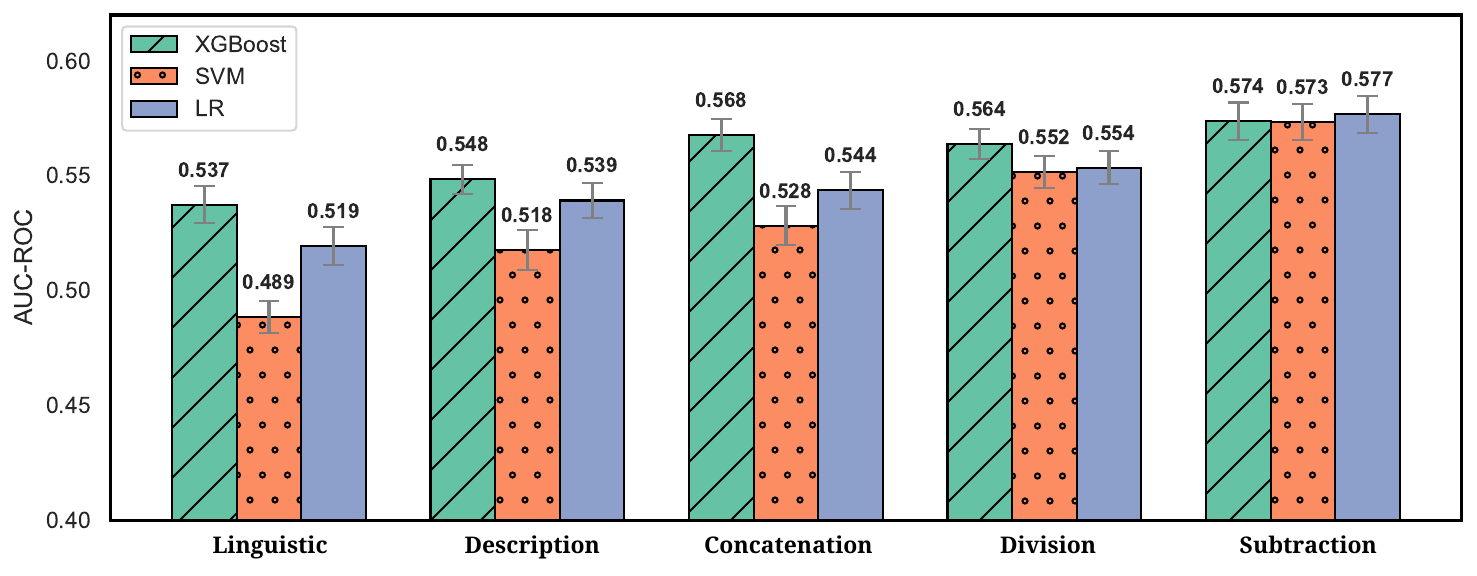}
    \caption{AUC-ROC performance of ML models using different aggregation techniques of \textbf{raw} NTP features.}
    \label{fig:Aggregation_results}
        \vspace{-0.8em}
\end{figure*}

We present the key results for the statistical NTP-based features in \autoref{fig:Engineered_Features_results} and the complete results in \autoref{tab:auc_performance} in \autoref{app:additional}. Results for the raw NTP-based features are shown in \autoref{fig:Aggregation_results}. Below, we discuss our main findings.

\textbf{Statistical features of NTPs can be competitive to VLM predictions} We begin by comparing the performance of statistical features derived from \emph{Description NTPs} with that of the $\mathbf{Pred}$ feature of \emph{LLaVA}. This comparison is natural, as the NTPs are extracted from the same model used for prediction. As shown in \autoref{fig:Engineered_Features_results}, \emph{LLaVA} $\mathbf{Pred}$ (dashed line) achieves slightly better performance than the statistical features extracted from the \emph{Description NTPs} (three leftmost bars), with the ROC AUC difference for LR being 0.013. Notice, however, that using \emph{LLaVA} $\mathbf{Pred}$ requires an additional forward pass of the VLM for every probe (and a single generated text can contain several probes). In contrast, \emph{Description NTP} features are obtained on-the-fly during generation and only require inference from a lightweight traditional ML model. 
Our results suggest that using \emph{Description NTPs} offers a compelling trade-off between performance and efficiency, making it a practical option for real-time applications where latency is paramount.

\textbf{Linguistic NTPs provide a modest improvement} We next examine whether incorporating statistical features from \emph{Linguistic NTPs} improves the performance of traditional ML models. Although using \emph{Linguistic NTPs} introduces additional computational costs compared to using only \emph{Description NTPs}, this cost remains relatively low. \emph{Linguistic NTPs} can be computed with a single forward pass of the language model after the text is generated, in contrast to the multiple VLM calls required for predictor VLMs (one for every probe).
As shown in \autoref{fig:Engineered_Features_results}, comparing the second group of bars (bars 4--6: \emph{Description} + \emph{Linguistic NTPs}) to the first group (bars 1--3: \emph{Description NTPs} only) reveals a consistent, albeit modest, performance gain across all ML models. The improvement in ROC AUC is approximately 0.01 and is not statistically significant, as indicated by overlapping confidence intervals. While these results suggest a positive effect from including \emph{Linguistic NTPs}, the benefit is limited, and further investigation is needed to understand their full potential.

\begin{figure*}[t]
\centering
\begin{subfigure}{0.4\linewidth}
  \centering
  \includegraphics[width=\linewidth]{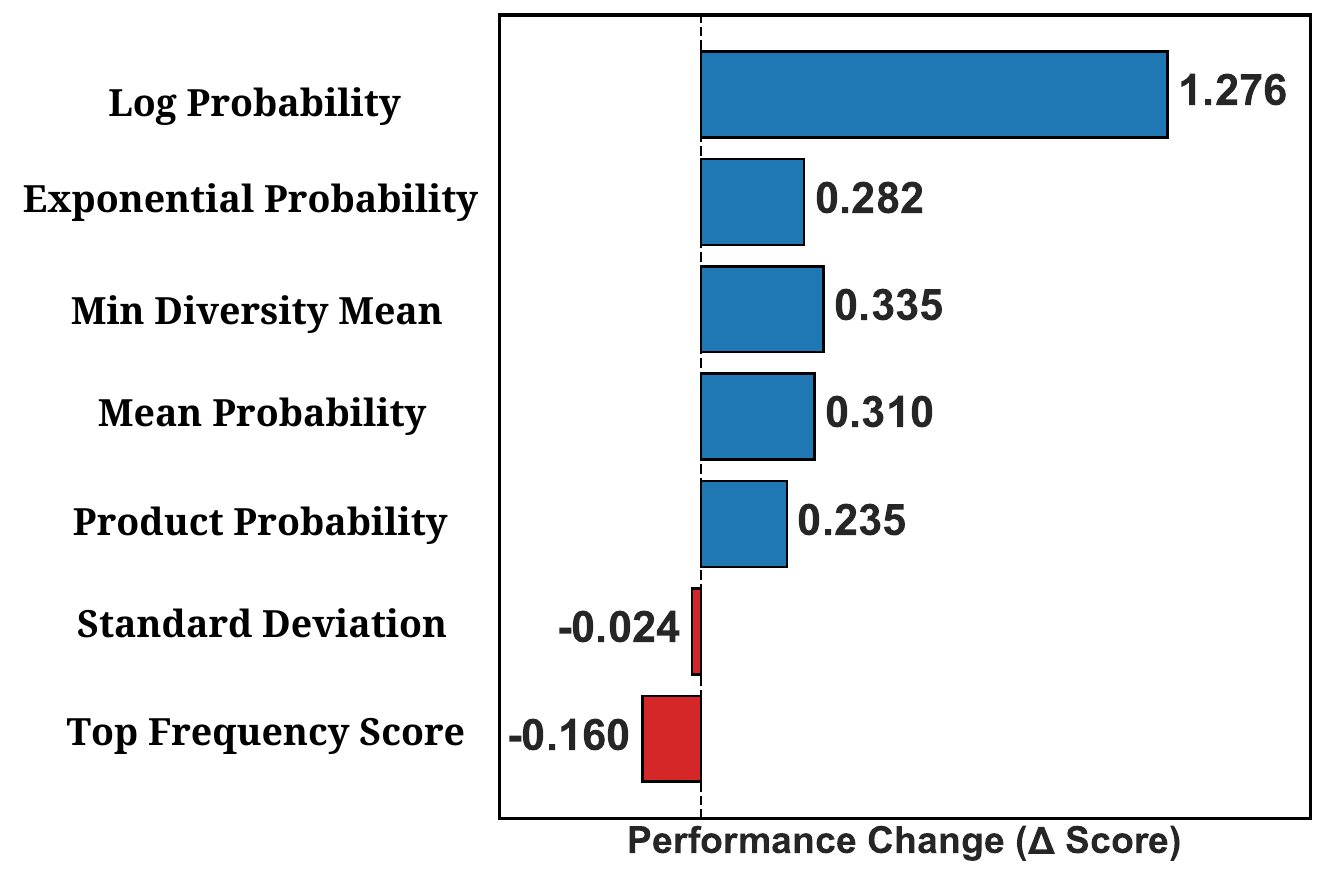}
\end{subfigure}%
\hfill
\begin{subfigure}{0.4\linewidth}
  \centering
  \includegraphics[width=\linewidth]{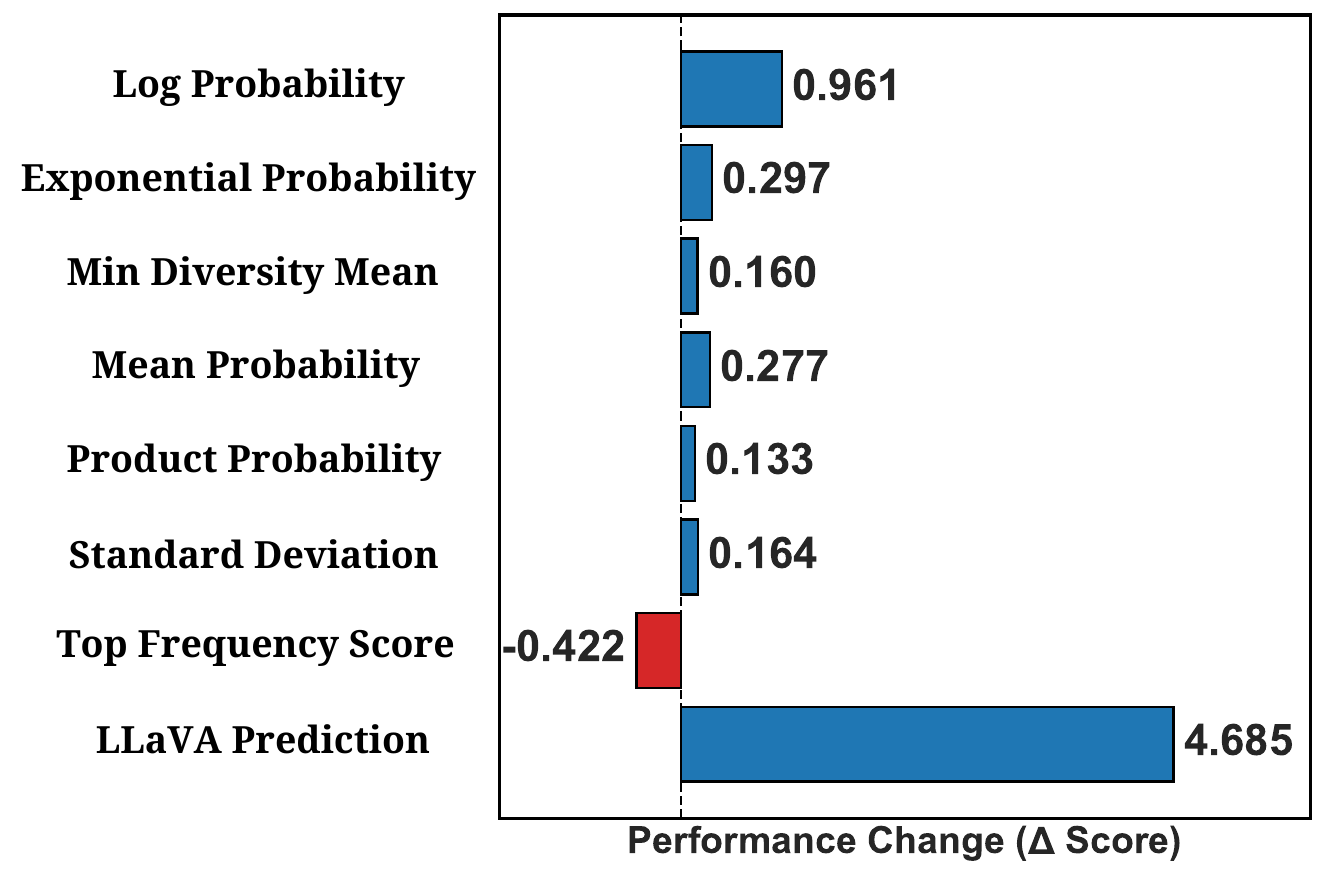}
\end{subfigure}
\caption{Leave-one-out ablation study on our features. Excluding (left) and including (right) \textit{LLaVA} predictions.}
\label{app:ablation_study_appendix}
\end{figure*}

\textbf{Statistical features of NTPs enhance VLM predictor performance.} So far, we have shown that NTP-based features offer a fast and lightweight solution for hallucination detection, although they moderately underperform compared to using the same VLM as a predictor. We now investigate whether combining both approaches can yield further improvements. As shown in \autoref{fig:Engineered_Features_results} (bars 7--9), augmenting the $\mathbf{Pred}$ feature with statistical features from \emph{Description NTPs} consistently improves performance across all traditional ML models. This indicates that NTPs alone can enhance hallucination detection when used alongside a predictor VLM. Specifically, the ROC AUC improvements over using \textit{LLaVA} $\mathbf{Pred}$ alone are 0.015, 0.028, 0.019 for XGBoost, SVM, and LR, respectively. 
We do not observe any further improvement regarding combining Linguistic NTP-based features (see bars 10--12). 

In addition to \emph{LLaVA}, we evaluate \emph{PaliGemma} as an alternative VLM predictor. While using an external predictor that differs from the generator introduces additional memory overhead, \emph{PaliGemma} $\mathbf{Pred}$ achieves substantially better performance than \emph{LLaVA} $\mathbf{Pred}$ (ROC AUC of 0.757 vs. 0.632).
We further assess whether combining both predictors improves performance. As shown in \autoref{fig:Engineered_Features_results} (bars 13--15), using both $\mathbf{Pred}$ features as input to SVM and LR yields an improvement over using \emph{PaliGemma} $\mathbf{Pred}$ alone, with an ROC AUC gain of 0.015. Finally, we examine whether adding statistical NTP-based features provides additional benefit in this combined predictor setup. While no improvement is observed for SVM and LR, XGBoost does show a performance gain when NTP features are included.



\textbf{Subtraction is the best aggregation of raw NTPs} Although our primary analysis emphasizes statistical features due to their superior performance compared to raw NTPs (compare bars 1--6 in \autoref{fig:Engineered_Features_results} to the bars in \autoref{fig:Aggregation_results}), we also explore raw NTP-based features, as they may offer additional insights for future work. In particular, we investigate how combining raw \emph{Description} and \emph{Linguistic NTPs} affects model performance.
As shown in \autoref{fig:Aggregation_results}, aggregation methods that aim to neutralize the influence of linguistic biases, such as element-wise subtraction or division of \emph{Description NTPs} by \emph{Linguistic NTPs}, consistently outperform simple concatenation across most ML models. Among these, subtraction yields the highest performance. This suggests that underlying linguistic patterns in the model shape the generated descriptions, and that these influences can be partially corrected through neutralization-based aggregation.

\subsection{Feature Importance Analysis}

We now assess the contribution of individual statistical features extracted from both \emph{Description} and \emph{Linguistic NTPs}. We consider multiple configurations, including models with and without the \textit{LLaVA} $\mathbf{Pred}$ feature. To evaluate feature importance, we conduct a leave-one-feature-out analysis: for each feature, we measure the change in performance ($\Delta$) as the difference in AUC-ROC between the full model (with all features) and the model with it removed. Results are presented in \autoref{app:ablation_study_appendix}. 

Unsurprisingly, the \textit{LLaVA} $\mathbf{Pred}$ feature is the most influential, providing a significantly larger performance gain than any of the NTP-based features. This aligns with its higher computational cost and the richer information it encapsulates from a full VLM inference pass. Among the NTP-based statistical features, we find that transformations of the probabilities, specifically, log-probabilities and exponentiated probabilities, are more informative than raw probabilities. This likely stems from the nature of the softmax distribution over generated tokens. These raw values offer limited variance and may obscure fine-grained differences in uncertainty. In contrast, applying logarithmic or exponential transformations expands the range, making subtle distinctions more detectable to the model.
Finally, time series features derived from the Discrete Fourier Transform (e.g., dominant frequencies) perform the worst. In some cases, including them even degrades model performance relative to the baseline, suggesting they may introduce noise or redundancy rather than useful signal.




\section{Conclusion}

In this paper, we explore the potential of leveraging uncertainty-related features to improve hallucination detection in text generated by VLMs. Specifically, we use NTPs extracted from VLMs in combination with traditional, efficient ML models to enhance detection performance while remaining computationally lightweight. Our results show that statistical features derived from \emph{Description NTPs} provide a lightweight and effective alternative to using VLM predictors.
While \emph{Linguistic NTPs} offer performance gains when $\mathbf{Pred}$ features are unavailable, they contribute little when such features are present, often making their additional computational cost unjustified. Finally, we find that combining NTP-based features with $\mathbf{Pred}$ scores leads to consistently improved detection performance, demonstrating their complementary nature.

We hope this work serves as a valuable resource for advancing the understanding and practical use of NTPs in hallucination detection. Our findings point to two promising directions for future research:
(1) developing efficient models of hallucination detection to support response refinement or the expression of uncertainty, and
(2) further investigating the relationship between \emph{Description} and \emph{Linguistic NTPs}, whose integration may prove valuable beyond hallucination detection.

\bibliography{custom}

\begin{thebibliography}{33}
\providecommand{\natexlab}[1]{#1}

\bibitem[{Benkirane et~al.(2024)Benkirane, Gongas, Pelles, Fuchs, Darmon, Stenetorp, Adelani, and S{\'a}nchez}]{benkirane2024machine}
Kenza Benkirane, Laura Gongas, Shahar Pelles, Naomi Fuchs, Joshua Darmon, Pontus Stenetorp, David~Ifeoluwa Adelani, and Eduardo S{\'a}nchez. 2024.
\newblock Machine translation hallucination detection for low and high resource languages using large language models.
\newblock \emph{arXiv preprint arXiv:2407.16470}.

\bibitem[{Beyer et~al.(2024)Beyer, Steiner, Pinto, Kolesnikov, Wang, Salz, Neumann, Alabdulmohsin, Tschannen, Bugliarello, Unterthiner, Keysers, Koppula, Liu, Grycner, Gritsenko, Houlsby, Kumar, Rong, Eisenschlos, Kabra, Bauer, Bosnjak, Chen, Minderer, Voigtlaender, Bica, Balazevic, Puigcerver, Papalampidi, Hénaff, Xiong, Soricut, Harmsen, and Zhai}]{PaliGemma}
Lucas Beyer, Andreas Steiner, André~Susano Pinto, Alexander Kolesnikov, Xiao Wang, Daniel Salz, Maxim Neumann, Ibrahim Alabdulmohsin, Michael Tschannen, Emanuele Bugliarello, Thomas Unterthiner, Daniel Keysers, Skanda Koppula, Fangyu Liu, Adam Grycner, Alexey~A. Gritsenko, Neil Houlsby, Manoj Kumar, Keran Rong, and 16 others. 2024.
\newblock \href {https://doi.org/10.48550/arXiv.2407.07726} {Paligemma: A versatile 3b vlm for transfer}.
\newblock \emph{CoRR}, abs/2407.07726.

\bibitem[{Bhadra et~al.(2021)Bhadra, Kelkar, Brooks, and Anastasio}]{bhadra2021hallucinations}
Sayantan Bhadra, Varun~A Kelkar, Frank~J Brooks, and Mark~A Anastasio. 2021.
\newblock On hallucinations in tomographic image reconstruction.
\newblock \emph{IEEE transactions on medical imaging}, 40(11):3249--3260.

\bibitem[{Chen et~al.(2024)Chen, Wang, Xue, Zhang, Yang, Li, and Chen}]{chen2024unified}
X.~Chen, C.~Wang, Y.~Xue, N.~Zhang, X.~Yang, Q.~Li, and H.~Chen. 2024.
\newblock Unified hallucination detection for multimodal large language models.
\newblock \emph{arXiv preprint}, arXiv:2402.03190.

\bibitem[{eun Cho and Maeng(2025)}]{cho2025influence}
Ye~eun Cho and Yunho Maeng. 2025.
\newblock \href {https://arxiv.org/abs/2502} {The influence of visual and linguistic cues on ignorance inference in vision-language models}.
\newblock \emph{arXiv e-prints}.

\bibitem[{Faggioli et~al.(2023)Faggioli, Dietz, Clarke, Demartini, Hagen, Hauff, Kando, Kanoulas, Potthast, Stein et~al.}]{faggioli2023perspectives}
Guglielmo Faggioli, Laura Dietz, Charles~LA Clarke, Gianluca Demartini, Matthias Hagen, Claudia Hauff, Noriko Kando, Evangelos Kanoulas, Martin Potthast, Benno Stein, and 1 others. 2023.
\newblock Perspectives on large language models for relevance judgment.
\newblock In \emph{Proceedings of the 2023 ACM SIGIR International Conference on Theory of Information Retrieval}, pages 39--50.

\bibitem[{Farquhar et~al.(2024)Farquhar, Kossen, Kuhn, and Gal}]{farquhar2024detecting}
Sebastian Farquhar, Jannik Kossen, Lorenz Kuhn, and Yarin Gal. 2024.
\newblock Detecting hallucinations in large language models using semantic entropy.
\newblock \emph{Nature}, 630(8017):625--630.

\bibitem[{Goldstein et~al.(2022)Goldstein, Zada, Buchnik, Schain, Price, Aubrey, Nastase, Feder, Emanuel, Cohen et~al.}]{goldstein2022shared}
Ariel Goldstein, Zaid Zada, Eliav Buchnik, Mariano Schain, Amy Price, Bobbi Aubrey, Samuel~A Nastase, Amir Feder, Dotan Emanuel, Alon Cohen, and 1 others. 2022.
\newblock Shared computational principles for language processing in humans and deep language models.
\newblock \emph{Nature neuroscience}, 25(3):369--380.

\bibitem[{Guan et~al.(2024)Guan, Liu, Wu, Xian, Li, Liu, Wang, Chen, Huang, Yacoob et~al.}]{guan2024hallusionbench}
Tianrui Guan, Fuxiao Liu, Xiyang Wu, Ruiqi Xian, Zongxia Li, Xiaoyu Liu, Xijun Wang, Lichang Chen, Furong Huang, Yaser Yacoob, and 1 others. 2024.
\newblock Hallusionbench: an advanced diagnostic suite for entangled language hallucination and visual illusion in large vision-language models.
\newblock In \emph{Proceedings of the IEEE/CVF Conference on Computer Vision and Pattern Recognition}, pages 14375--14385.

\bibitem[{Guerreiro et~al.(2022)Guerreiro, Voita, and Martins}]{guerreiro2022looking}
Nuno~M Guerreiro, Elena Voita, and Andr{\'e}~FT Martins. 2022.
\newblock Looking for a needle in a haystack: A comprehensive study of hallucinations in neural machine translation.
\newblock \emph{arXiv preprint arXiv:2208.05309}.

\bibitem[{Gunjal et~al.(2024)Gunjal, Yin, and Bas}]{gunjal2024detecting}
Anisha Gunjal, Jihan Yin, and Erhan Bas. 2024.
\newblock Detecting and preventing hallucinations in large vision language models.
\newblock In \emph{Proceedings of the AAAI Conference on Artificial Intelligence}, volume~38, pages 18135--18143.

\bibitem[{Leng et~al.(2024)Leng, Zhang, Chen, Li, Lu, Miao, and Bing}]{leng2024mitigating}
Sicong Leng, Hang Zhang, Guanzheng Chen, Xin Li, Shijian Lu, Chunyan Miao, and Lidong Bing. 2024.
\newblock Mitigating object hallucinations in large vision-language models through visual contrastive decoding.
\newblock In \emph{Proceedings of the IEEE/CVF Conference on Computer Vision and Pattern Recognition}, pages 13872--13882.

\bibitem[{Li et~al.(2019)Li, Yatskar, Yin, Hsieh, and Chang}]{li2019visualbertsimpleperformantbaseline}
Liunian~Harold Li, Mark Yatskar, Da~Yin, Cho-Jui Hsieh, and Kai-Wei Chang. 2019.
\newblock \href {https://arxiv.org/abs/1908.03557} {Visualbert: A simple and performant baseline for vision and language}.
\newblock \emph{Preprint}, arXiv:1908.03557.

\bibitem[{Li et~al.(2024)Li, Lyu, Geng, Zhu, Panov, and Karray}]{li2024reference}
Qing Li, Chenyang Lyu, Jiahui Geng, Derui Zhu, Maxim Panov, and Fakhri Karray. 2024.
\newblock Reference-free hallucination detection for large vision-language models.
\newblock \emph{arXiv preprint}, arXiv:2408.05767.

\bibitem[{Liu et~al.(2024)Liu, Li, Wu, and Lee}]{liu2024visual}
Haotian Liu, Chunyuan Li, Qingyang Wu, and Yong~Jae Lee. 2024.
\newblock Visual instruction tuning.
\newblock \emph{Advances in neural information processing systems}, 36.

\bibitem[{Liu et~al.(2023)Liu, {\v{S}}krjanec, and Demberg}]{Liu2023TemperaturescalingSE}
Tong Liu, Iza {\v{S}}krjanec, and Vera Demberg. 2023.
\newblock \href {https://api.semanticscholar.org/CorpusID:265220963} {Temperature-scaling surprisal estimates improve fit to human reading times - but does it do so for the "right reasons"?}
\newblock In \emph{Annual Meeting of the Association for Computational Linguistics}.

\bibitem[{Lu et~al.(2021)Lu, Zhang, Zheng, and Li}]{lu2021communication}
Jiahui Lu, Meishan Zhang, Yan Zheng, and Qiyu Li. 2021.
\newblock \href {https://doi.org/10.3390/ijerph182211933} {Communication of uncertainty about preliminary evidence and the spread of its inferred misinformation during the covid-19 pandemic—a weibo case study}.
\newblock \emph{International Journal of Environmental Research and Public Health}, 18(22):11933.

\bibitem[{Luo et~al.(2024)Luo, Cao, Lee, Johnson, and Lee}]{luo2024vvlm}
Tiange Luo, Ang Cao, Gunhee Lee, Justin Johnson, and Honglak Lee. 2024.
\newblock vvlm: Exploring visual reasoning in vlms against language priors.
\newblock \emph{OpenReview lCqNxBGPp5}.

\bibitem[{Ma et~al.(2025)Ma, Chen, Wang, and Zhang}]{ma2025estimating}
Huan Ma, Jingdong Chen, Guangyu Wang, and Changqing Zhang. 2025.
\newblock Estimating llm uncertainty with logits.
\newblock \emph{arXiv preprint arXiv:2502.00290}.

\bibitem[{Magesh et~al.(2024)Magesh, Surani, Dahl, Suzgun, Manning, and Ho}]{magesh2024hallucination}
Varun Magesh, Faiz Surani, Matthew Dahl, Mirac Suzgun, Christopher~D Manning, and Daniel~E Ho. 2024.
\newblock Hallucination-free? assessing the reliability of leading ai legal research tools.
\newblock \emph{arXiv preprint arXiv:2405.20362}.

\bibitem[{Maleki et~al.(2024)Maleki, Padmanabhan, and Dutta}]{maleki2024ai}
Negar Maleki, Balaji Padmanabhan, and Kaushik Dutta. 2024.
\newblock Ai hallucinations: a misnomer worth clarifying.
\newblock In \emph{2024 IEEE conference on artificial intelligence (CAI)}, pages 133--138. IEEE.

\bibitem[{Quevedo et~al.(2024)Quevedo, Yero, Koerner, Rivas, and Cerny}]{quevedo2024detecting}
E.~Quevedo, J.~Yero, R.~Koerner, P.~Rivas, and T.~Cerny. 2024.
\newblock Detecting hallucinations in large language model generation: A token probability approach.
\newblock \emph{arXiv preprint}, arXiv:2405.19648.

\bibitem[{Radford et~al.(2021)Radford, Kim, Hallacy, Ramesh, Goh, Agarwal, Sastry, Askell, Mishkin, Clark, Krueger, and Sutskever}]{radford2021learningtransferablevisualmodels}
Alec Radford, Jong~Wook Kim, Chris Hallacy, Aditya Ramesh, Gabriel Goh, Sandhini Agarwal, Girish Sastry, Amanda Askell, Pamela Mishkin, Jack Clark, Gretchen Krueger, and Ilya Sutskever. 2021.
\newblock \href {https://arxiv.org/abs/2103.00020} {Learning transferable visual models from natural language supervision}.
\newblock \emph{Preprint}, arXiv:2103.00020.

\bibitem[{Sarkar(2024)}]{sarkar2024large}
Advait Sarkar. 2024.
\newblock Large language models cannot explain themselves.
\newblock \emph{arXiv preprint arXiv:2405.04382}.

\bibitem[{Shrivastava et~al.(2023)Shrivastava, Liang, and Kumar}]{Shrivastava2023LlamasKW}
Vaishnavi Shrivastava, Percy Liang, and Ananya Kumar. 2023.
\newblock \href {https://api.semanticscholar.org/CorpusID:265213392} {Llamas know what gpts don't show: Surrogate models for confidence estimation}.
\newblock \emph{ArXiv}, abs/2311.08877.

\bibitem[{Simhi et~al.(2025)Simhi, Itzhak, Barez, Stanovsky, and Belinkov}]{simhi2025trust}
Adi Simhi, Itay Itzhak, Fazl Barez, Gabriel Stanovsky, and Yonatan Belinkov. 2025.
\newblock Trust me, i'm wrong: High-certainty hallucinations in llms.
\newblock \emph{arXiv preprint arXiv:2502.12964}.

\bibitem[{Tang et~al.(2025)Tang, Chatterjee, and Garg}]{tang2025mitigatinghallucinatedtranslationslarge}
Zilu Tang, Rajen Chatterjee, and Sarthak Garg. 2025.
\newblock \href {https://arxiv.org/abs/2501.17295} {Mitigating hallucinated translations in large language models with hallucination-focused preference optimization}.
\newblock \emph{Preprint}, arXiv:2501.17295.

\bibitem[{Wang et~al.(2024)}]{wang2024linguistic}
Fei Wang and 1 others. 2024.
\newblock Can linguistic knowledge improve multimodal alignment in vision-language pretraining?
\newblock \emph{ACM Transactions on Multimedia Computing, Communications and Applications}, 20(12):1--22.

\bibitem[{Wu et~al.(2024)Wu, Guan, Li, Huang, Liu, Wang, Xian, Shrivastava, Huang, Boyd-Graber et~al.}]{wu2024autohallusion}
Xiyang Wu, Tianrui Guan, Dianqi Li, Shuaiyi Huang, Xiaoyu Liu, Xijun Wang, Ruiqi Xian, Abhinav Shrivastava, Furong Huang, Jordan~Lee Boyd-Graber, and 1 others. 2024.
\newblock Autohallusion: Automatic generation of hallucination benchmarks for vision-language models.
\newblock \emph{arXiv preprint arXiv:2406.10900}.

\bibitem[{Xu et~al.(2024)Xu, Jain, and Kankanhalli}]{xu2024hallucination}
Ziwei Xu, Sanjay Jain, and Mohan Kankanhalli. 2024.
\newblock Hallucination is inevitable: An innate limitation of large language models.
\newblock \emph{arXiv preprint arXiv:2401.11817}.

\bibitem[{Zhao et~al.(2024{\natexlab{a}})Zhao, Chen, Yang, Liu, Deng, Cai, Wang, Yin, and Du}]{zhao2024explainability}
Haiyan Zhao, Hanjie Chen, Fan Yang, Ninghao Liu, Huiqi Deng, Hengyi Cai, Shuaiqiang Wang, Dawei Yin, and Mengnan Du. 2024{\natexlab{a}}.
\newblock Explainability for large language models: A survey.
\newblock \emph{ACM Transactions on Intelligent Systems and Technology}, 15(2):1--38.

\bibitem[{Zhao et~al.(2024{\natexlab{b}})Zhao, Yuan, Wei, Zhang, Gu, Ran, Wang, Wu, Zhang, Zhang, and Shou}]{zhao2024evolvedirector}
Rui Zhao, Hangjie Yuan, Yujie Wei, Shiwei Zhang, Yuchao Gu, Lingmin Ran, Xiang Wang, Jay~Zhangjie Wu, David~Junhao Zhang, Yingya Zhang, and Mike~Zheng Shou. 2024{\natexlab{b}}.
\newblock \href {https://openreview.net/forum?id=28bFUt6rUY} {Evolvedirector: Approaching advanced text-to-image generation with large vision-language models}.
\newblock In \emph{The Thirty-eighth Annual Conference on Neural Information Processing Systems}.

\bibitem[{Zhu et~al.(2024)Zhu, Ji, Chen, Xu, Ye, and Liu}]{zhu2024ibd}
Lanyun Zhu, Deyi Ji, Tianrun Chen, Peng Xu, Jieping Ye, and Jun Liu. 2024.
\newblock Ibd: Alleviating hallucinations in large vision-language models via image-biased decoding.
\newblock \emph{arXiv preprint arXiv:2402.18476}.

\end{thebibliography}

\newpage
\appendix
\onecolumn

\section*{Appendix}
\section{Data Collection Pipeline}
\label{sec: AppendixPipeline}
\begin{figure}[h!]
    \begin{center}
        \includegraphics[width=0.9\textwidth]{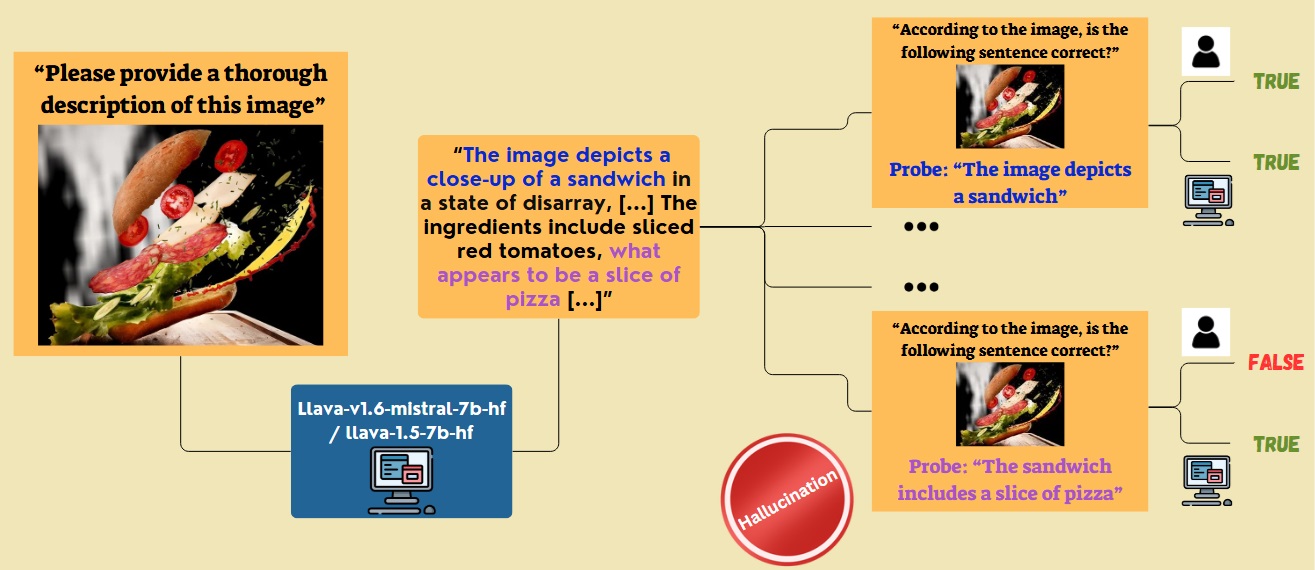}
    \end{center}
    \caption{An illustration of the data collection pipeline.}
    \label{fig:pipeline}
\end{figure}

In this section, we will demonstrate the data collection pipeline and the calls for the LLM for a single example of an image. In Figure \ref{fig:pipeline} the pipeline starts by instructing the LLM to return a description of the image, which it does. The description it generates in the figure contains a hallucination which is marked in  - marked in \textbf{\textcolor{custompurple}{purple}}. In \textbf{\textcolor{customblue}{blue}} there is a correct statement though. Four probes are manually derived from this generated description, and the model is asked whether each probe is correct or not. This is judged by human feedback (represented by the person's icon), which represents the \say{true labels}, and by the \emph{LLaVA} model (represented by the computer's icon). In the first probe both the model and the human judgments are the same, and they both agree on the correctness of the probe. This is not the case with the fourth probe which is a false statement the model generated, but the model predicts it is correct. From this two calls for the \emph{LLaVA} model, we can collect all features mentioned in \S\ref{sec:data}, and the other features which were not mentioned in this paper.

\section{NTPs analysis}
\label{appendix:motivation}
In order to justify the use of both the \emph{Description NTPs} and \emph{Linguistic NTPs}, some statistics were examined of both types.

\subsection{Description NTPs}

\begin{figure*}
\centering
\begin{subfigure}{.5\textwidth}
  \centering
  \includegraphics[width=\textwidth]{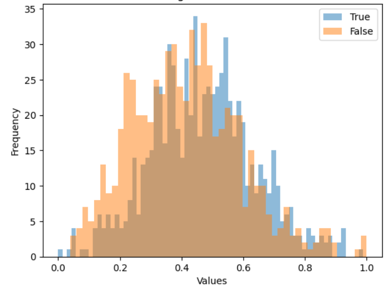}
\end{subfigure}%
\begin{subfigure}{.5\textwidth}
  \centering
  \includegraphics[width=.85\textwidth]{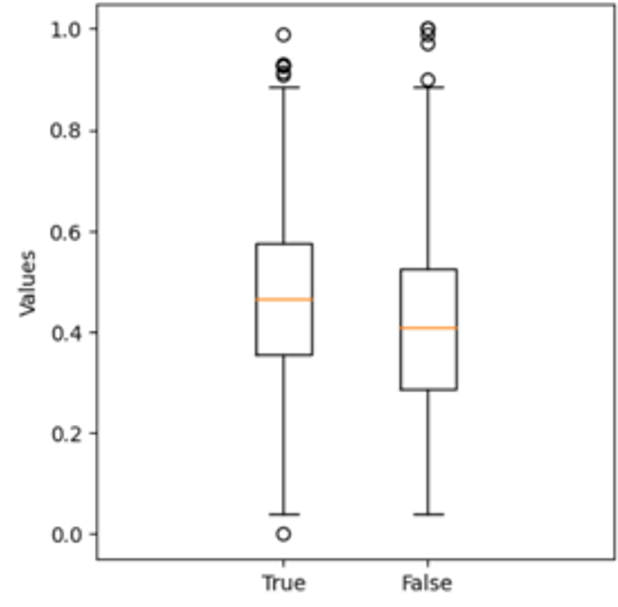}
\end{subfigure}
\caption{Distributions (left) and box-plot (right) of \emph{Description NTPs} in contexts that do not contain hallucinations and in contexts that do. In the box plot, the left box corresponds to the \emph{Description NTPs} in contexts that do not contain hallucinations, and the right one corresponds \emph{Description NTPs} in contexts that contain hallucination(s). In both plots, NTPs are aggregated using a geometric mean to produce a single number for each context.}
\label{fig:NTPs}
\end{figure*}

Figure  \ref{fig:NTPs} demonstrates that the \emph{Description NTPs} are a viable feature that can differentiate in some manner between texts that do not contain hallucinations and texts which do. Though the distributions share a great amount of probability mass, the difference between these two distributions is still notable, and the difference between the two can also be observed in the box plot. Hence, we believe in the potential of these NTPs as a useful feature that can assist in detecting hallucinations.

\subsection{Linguistic NTPs}
\begin{figure*}[t]
\centering
\begin{subfigure}{.5\textwidth}
  \centering
  \includegraphics[width=\textwidth]{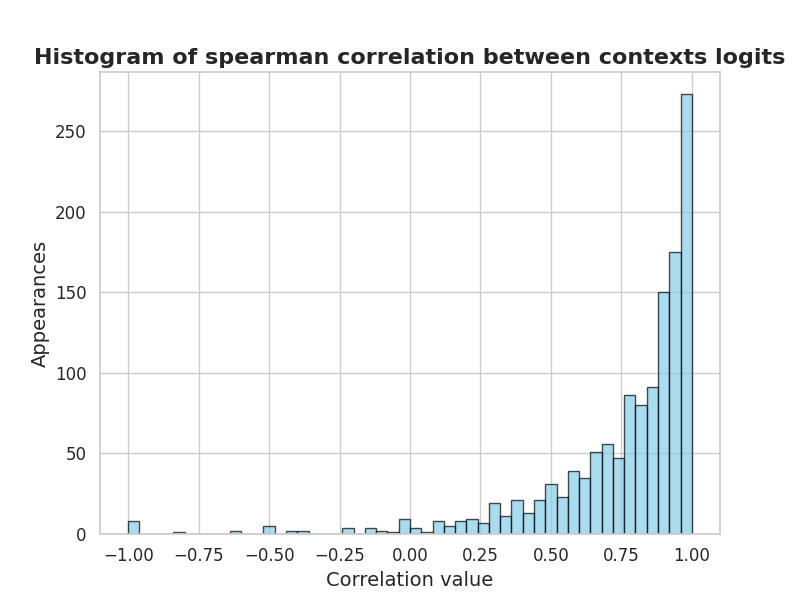}
\end{subfigure}%
\begin{subfigure}{.5\textwidth}
  \centering
  \includegraphics[width=.95\textwidth]{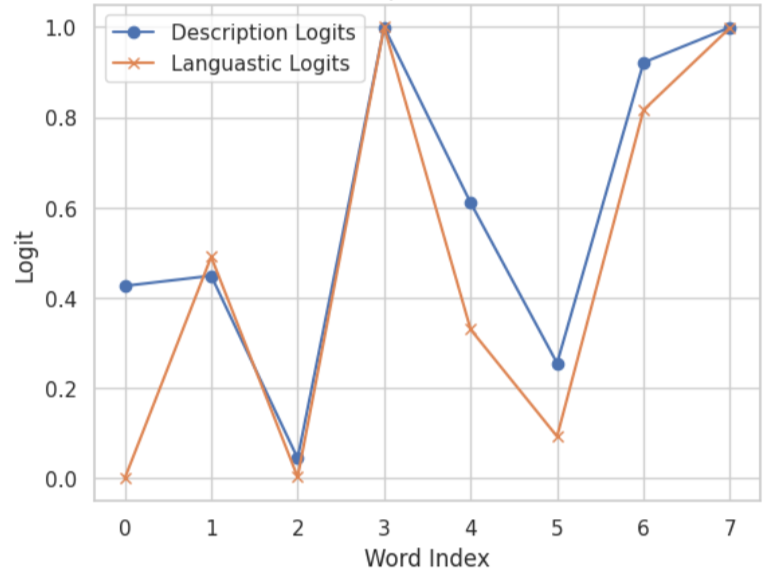}
\end{subfigure}
\caption{Histogram of the Spearman Correlation values between the \emph{Description NTPs} and the \emph{Linguistic NTPs} (left). A single sampled example of the similar trends both NTPs exhibit in one of the contexts (right)}
\label{fig:linguistic}
\end{figure*}

We witnessed the merits of using the \emph{Description NTPs} for detecting hallucinations, and their analysis revealed some repetitive peaks and patterns, which were hypothesized to be connected to the linguistic component of the NTPs. To examine the influence of using the collected \emph{Linguistic NTPs}, as a proxy for the linguistic part of the text, we first checked the correlation between both types of NTPs. It was hypothesized that a high correlation between them can indicate the merits of using \emph{Linguistic NTPs} as a tool to decrease the noise and anomalies coming from the linguistic part of the generation. Considering the Spearman Correlation, the result was that the average correlation is $0.755$, and the median correlation was $0.857$. Figure \ref{fig:linguistic} illustrates the distribution of correlations among the different contexts, and demonstrates the strong correlation between both NTPs types.

\section{Hyperparameter Tuning}
\label{app:hyperparams}

For our three ML models, we performed hyperparameter tuning using grid search to identify the optimal parameters that maximize the AUC-ROC score on the validation set. The best-performing parameters were then used to train the final model, which was evaluated on the test set.

LR and SVM were implemented using the \texttt{LogisticRegression} and \texttt{SVC} classes from the \texttt{scikit-learn} library. The XGBoost model was implemented using the \texttt{train} function from the \texttt{xgboost} library. The specific search grids for each model are detailed below.

\textbf{LR:} We optimized the regularization strength and penalty type while considering different solvers. The search grid included:
\begin{itemize}
    \item $C \in \{0.1, 1, 10, 100\}$ (Regularization strength)
    \item Penalty type: $\{L_1,L_2\}$
    \item Solvers: $\{\text{lbfgs, liblinear, newton-cg, newton-cholesky, sag, saga}\}$
\end{itemize}

\textbf{SVM:} We explored different values for the regularization parameter ($C$), kernel type, and kernel coefficient ($\gamma$) for the rbf kernel:
\begin{itemize}
    \item $C \in \{0.1, 1, 10, 100\}$
    \item Kernel type: $\{\text{linear, rbf}\}$
    \item $\gamma \in \{\text{scale, auto, 1, 0.1, 0.01, 0.001}\}$
\end{itemize}

\textbf{XGBoost:} We tuned multiple hyperparameters including tree depth, learning rate, regularization terms, and subsampling ratios:
\begin{itemize}
    \item Maximum tree depth: $\{3, 5\}$
    \item Learning rate: $\{0.1, 0.2\}$
    \item Minimum child weight: $\{3, 5, 7\}$
    \item Gamma (regularization parameter): $\{0.01, 0.1\}$
    \item Subsample ratio: $\{0.6, 0.7\}$
    \item Column sampling ratio: $\{0.6, 0.7\}$
    \item L1 regularization ($\alpha$): $\{0.1, 1, 10\}$
    \item L2 regularization ($\lambda$): $\{1, 10, 100\}$
\end{itemize}

Grid search with cross-validation was employed to systematically evaluate all parameter combinations. The best-performing hyperparameter set for each model was then used for final training and evaluation on the test dataset.

\newpage

\section{Tabular Results for Figure \ref{fig:Engineered_Features_results}}
\label{app:additional}

\begin{table}[h]
    \centering
    \begin{tabular}{ccccc}
        \midrule
        \multicolumn{5}{c}{\textbf{ML Models Performance}} \\
        \toprule
        \textbf{Preds} & \textbf{Linguistic} & \textbf{XGBoost} & \textbf{SVM} & \textbf{LR} \\  
        \midrule
        \multirow{2}{*}{No Preds} & No & 0.589 $\pm$ 0.008 & 0.597 $\pm$ 0.008 & 0.606 $\pm$ 0.007 \\  
                                   & Yes    & 0.599 $\pm$ 0.008 & 0.611 $\pm$ 0.008 & 0.615 $\pm$ 0.007 \\  
        \midrule
        \multirow{2}{*}{\emph{LLaVA}}    & No & 0.647 $\pm$ 0.007 & 0.660 $\pm$ 0.008 & 0.651 $\pm$ 0.007 \\  
                                   & Yes    & 0.649 $\pm$ 0.008 & 0.658 $\pm$ 0.008 & 0.652 $\pm$ 0.007 \\  
        \midrule
        \multirow{2}{*}{\emph{PaliGemma}} & No & 0.739 $\pm$ 0.006 & 0.758 $\pm$ 0.005 & 0.761 $\pm$ 0.006 \\  
                                   & Yes    & 0.735 $\pm$ 0.007 & 0.759 $\pm$ 0.006 & 0.761 $\pm$ 0.006 \\  
        \midrule
        \multirow{2}{*}{\emph{LLaVA} and \emph{PaliGemma}} & No & 0.760 $\pm$ 0.006 & 0.772 $\pm$ 0.005 & 0.771 $\pm$ 0.005 \\  
                                             & Yes    & 0.761 $\pm$ 0.007 & 0.770 $\pm$ 0.006 & 0.769 $\pm$ 0.005 \\ 
        \midrule
        \multicolumn{5}{c}{\textbf{VLM Performance}} \\
        \midrule
        \textbf{VLM Type} & \textbf{Raw Score} & \textbf{XGBoost} & \textbf{SVM} & \textbf{LR} \\ 
        \midrule
        \emph{LLaVA}     & 0.632 $\pm$ 0.007 & -- & -- & -- \\  
        \emph{PaliGemma} & 0.757 $\pm$ 0.005 & -- & -- & -- \\  
        \emph{LLaVA} and \emph{PaliGemma}             & -- & 0.754 $\pm$ 0.006 & 0.772 $\pm$ 0.005 & 0.772 $\pm$ 0.005 \\
        \bottomrule
    \end{tabular}
    \caption{Detailed AUC-ROC performance (with 95\% confidence intervals) of traditional ML models and VLMs across different configurations. The upper section evaluates ML models using only NTP-based features as distinct inputs or in combination with VLM predictions. The lower section reports standalone VLM performance. Where a single VLM prediction is directly adopted as the final prediction and when both VLM predictions are combined, ML models utilize both prediction features to make the final prediction. }
    \label{tab:auc_performance}
\end{table}

\end{document}